\documentclass[letterpaper]{article} % DO NOT CHANGE THIS
\usepackage{aaai25}  % DO NOT CHANGE THIS
\usepackage{times}  % DO NOT CHANGE THIS
\usepackage{helvet}  % DO NOT CHANGE THIS
\usepackage{courier}  % DO NOT CHANGE THIS
\usepackage[hyphens]{url}  % DO NOT CHANGE THIS
\usepackage{graphicx} % DO NOT CHANGE THIS
\usepackage{natbib}  % DO NOT CHANGE THIS AND DO NOT ADD ANY OPTIONS TO IT
\usepackage{caption} % DO NOT CHANGE THIS AND DO NOT ADD ANY OPTIONS TO IT
\usepackage{algorithm}
\usepackage{algorithmic}

\usepackage{newfloat}
\usepackage{listings}
\DeclareCaptionStyle{ruled}{labelfont=normalfont,labelsep=colon,strut=off} % DO NOT CHANGE THIS
\floatstyle{ruled}
\newfloat{listing}{tb}{lst}{}
\floatname{listing}{Listing}
\usepackage{multirow}
\usepackage{bbm}
\newcommand{\one}[1]{\mathbbm{1}_{[#1]}}
\usepackage{soul}

\usepackage[table,xcdraw]{xcolor}

\usepackage{xcolor,colortbl}
\definecolor{ffe1da}{RGB}{255,225,218}
\definecolor{F7E0D5}{RGB}{247,224,213}
\definecolor{darkF7E0D5}{RGB}{209,154,128}
\usepackage{wrapfig}
\usepackage{caption}
\usepackage{booktabs}
\usepackage{amsmath}  
\usepackage{amssymb}  

\usepackage{cuted}
\usepackage{capt-of}
\usepackage{soul}

\title{OpenVIS: Open-vocabulary Video Instance Segmentation}
\author {
    Pinxue Guo\textsuperscript{\rm 1},
    Tony Huang\textsuperscript{\rm 2},
    Peiyang He\textsuperscript{\rm 2},
    Xuefeng Liu\textsuperscript{\rm 2},
    Tianjun Xiao\textsuperscript{\rm 2},\\
    Zhaoyu Chen\textsuperscript{\rm 1},
    Wenqiang Zhang\textsuperscript{\rm 1}
}
\affiliations {
    \textsuperscript{\rm 1}Academy for Engineering and Technology, Fudan University\\
    \textsuperscript{\rm 2}Amazon Web Services\\
}

\usepackage{bibentry}
\begin{document}

\maketitle

\begin{abstract}
Open-vocabulary Video Instance Segmentation (OpenVIS) can simultaneously detect, segment, and track arbitrary object categories in a video, without being constrained to categories seen during training. 
In this work, we propose InstFormer, a carefully designed framework for the OpenVIS task that achieves powerful open-vocabulary capabilities through lightweight fine-tuning with limited-category data. InstFormer begins with the open-world mask proposal network, encouraged to propose all potential instance class-agnostic masks by the contrastive instance margin loss. Next, we introduce InstCLIP, adapted from pre-trained CLIP with Instance Guidance Attention, which encodes open-vocabulary instance tokens efficiently. These instance tokens not only enable open-vocabulary classification but also offer strong universal tracking capabilities. Furthermore, to prevent the tracking module from being constrained by the training data with limited categories, we propose the universal rollout association, which transforms the tracking problem into predicting the next frame’s instance tracking token.
The experimental results demonstrate the proposed InstFormer achieve state-of-the-art capabilities on a comprehensive OpenVIS  evaluation benchmark, while also achieves competitive performance in fully supervised VIS task.
\end{abstract}

\vspace{-4mm}

\section{Introduction}

Video understanding~\cite{pmlr-v139-bertasius21a, hong2022lvos, guo2022adaptive, wang2022dpcnet, zhou2023memory, li2023videochat} is a challenging yet significant computer vision task that requires specialized algorithms and techniques, surpassing the difficulty of image understanding. To achieve a more thorough understanding, Video Instance Segmentation (VIS)~\cite{yang2019video} has been proposed, which can simultaneously detect, segment, and track instances in a given video, becoming a new research hotspot. 
Despite significant progress, current VIS models possess an inherent limitation. They can only segment objects within the boundaries of their training data, meaning they are unable to identify objects beyond the categories present in the training set. Consequently, their video understanding remains restricted. Moreover, identifying new categories requires retraining with additional annotated data, leading to substantial time and resource investment.
To address the limitation, we investigate a novel computer vision task called Open-vocabulary Video Instance Segmentation (OpenVIS). This task focuses on detecting, segmenting, and tracking instances in videos based on the category names of target objects, regardless of whether those categories have been seen during the training stage. 

\begin{figure}[t]
    \centering
    \includegraphics[width=1.0\linewidth]{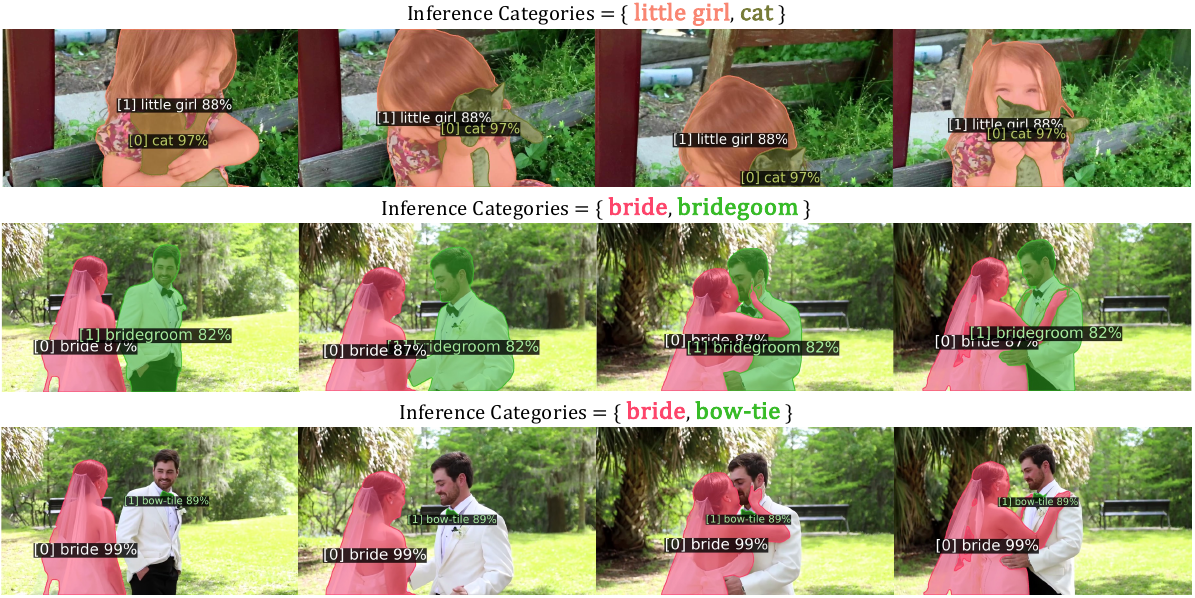}
    \vspace{-5mm}
    \caption{\textbf{Visualization of Open-vocabulary Video Instance Segmentation.} OpenVIS simultaneously segments, detects, and tracks arbitrary objects in a video according to their corresponding text description. The proposed InstFormer can accurately identify various objects based on their respective category names in a video, irrespective of whether the category is included in the training set.}
    \label{fig:teaser}
    \vspace{-7mm}
\end{figure}

Although recent pre-trained Vision-Language Models (VLMs)~\cite{radford2021learning,yao2021filip} have shown promising results in zero-shot classification and provide good foundation for open-vocabulary video instance segmentation, significant challenges still remain in leveraging these static, image-level VLMs for this video, instance-level task.
To eliminate this gap, we propose InstFormer, a carefully designed framework tailored for the OpenVIS task that achieves robust open-vocabulary capabilities through lightweight fine-tuning on a limited-category labeled dataset.
\textbf{Firstly}, since any object, rather than fixed categories, might be selected for identification by the end user, InstFormer first performs the open-world mask proposal by incorporating a margin instance contrastive loss into a query-based mask proposal network to generates class-agnostic instance masks, with the goal of proposing as many distinct instances within a given video as possible to meet the flexible needs of open-world perception.
\textbf{Secondly}, obtaining open-vocabulary representations of instances to enable classification and tracking across frames is non-trivial. 
Leveraging the zero-shot capabilities of pre-trained VLMs like CLIP~\cite{radford2021learning} by directly inputting masked instance images is suboptimal and inefficient for video tasks due to the domain gap between pre-training on natural images and testing on masked images, as well as the need to run the VLM’s vision encoder multiple times per frame. So we propose the InstCLIP, a variant of CLIP adapted with the proposed Instance Guidance Attention, which directs instance tokens to attend different instance regions simultaneously by the generated guidance according to multiple mask proposals. These instance tokens not only enable open-vocabulary classification but also offer strong universal tracking capabilities. 
\textbf{Thirdly}, the training process of current instance trackers with fixed-category datasets in video instance segmentation presents a significant challenge when it comes to tracking open-vocabulary instances. To address this issue, we propose Universal Rollout Association, where the rollout tracker is trained to predict instance tokens of the next frame to achieve tracking. The rollout tracker is implemented with a simple yet history-aware RNN layer, predicting instance tokens for the next frame based on previous tracking tokens. This prediction training is independent of categories, enabling the rollout tracker to handle open-vocabulary task.

To facilitate research on this novel task, we propose an evaluation benchmark that utilizes readily available datasets to thoroughly assess the performance. In our benchmark, the OpenVIS model will be trained with a limited number of categories, and subsequently tested on a large number of categories. Specifically, we evaluate the proposed model on YouTube-VIS, BURST, LVVIS, and UVO datasets, encompassing a large number of novel categories, to comprehensively assess its diverse capacities. However, the training process only see the data of YouTube-VIS, which comprises only 40 categories.
The experimental results demonstrate the proposed InstFormer achieves state-of-the-art capabilities in OpenVIS and competitive performance in fully supervised VIS. This indicates that InstFormer retains most of VLM’s zero-shot capabilities while optimizing for specific domains, providing a sound solution for scenarios needing both extreme domain performance and generalization.
Our contributions can be summarized as follows:

\begin{itemize}
    \item We propose the InstFormer framework, which achieves open-vocabulary capabilities through lightweight fine-tuning on limited-category data, to explore the novel OpenVIS task and introduce a comprehensive evaluation benchmark.
    \item We introduce the contrastive instance margin loss to open-world mask proposal network to encourage the generation of distinct instance proposals.
    \item We present InstCLIP, designed to embed each open-vocabulary instance with an instance token. The resulting instance tokens not only enable efficient open-vocabulary classification for multiple instances but also prove effective in subsequent open-vocabulary instance tracking.
    \item We propose universal rollout association, which achieves tracking by training the tracker to predict instance tokens of the next frame, to overcome the limitations of trackers trained on fixed-category data that struggle to generalize to open-vocabulary instances.
\end{itemize}

\section{Related Work}
\subsection{Video Instance Segmentation}
 There are two main prevailing paradigms for Video Instance Segmentation~\cite{yang2019video}:
 offline approaches and online approaches. 
 Offline approaches such as VisTR~\cite{wang2021end}, Mask2Former-VIS~\cite{cheng2021mask2former}, and SeqFormer~\cite{wu2021seqformer}, take the whole video as input and utilize instance queries to predict the instance sequence of the entire video or clip in a single step. Despite the high performance on popular datasets, the requirement of the whole video limits the application of offline methods, especially for long video and on-going video scenarios.
 Online approaches like MaskTrack RCNN~\cite{yang2019video}, MaskProp~\cite{bertasius2020classifying}, MinVIS~\cite{huang2022minvis}, IDOL~\cite{wu2022idol}, and DVIS~\cite{zhang2023dvis} independently process each frame to obtain all the mask proposals and their corresponding categories and then track these instance proposals by post-processing methods. 
Our approach also builds on this way which provides us with more flexibility and convenience in predicting potential mask proposals and using InstCLIP for instance classification.

\subsection{Vision-Language Models}
Vision-language models aim to bridge the gap between visual and textual modalities, gaining significant attention due to their impressive performance in visual representation learning. Recently, benefiting from the large-scale training data, pre-trained VLMs like CLIP~\cite{radford2021learning} and FLIP~\cite{yao2021filip} have further shown strong zero-shot object recognition capability. For example, after training on 400 million image-text paired data, CLIP (VIT-L as visual encoder) can achieve 76.2\% zero-shot accuracy on ImageNet, encouraging a series of downstream vision tasks to involving these pre-trained models, such as classification~\cite{radford2021learning, huang2022unsupervised}, captioning~\cite{hu2022scaling}, retrieval~\cite{liu2021image}, and segmentation~\cite{xu2021simple}.
Despite VLMs enabling powerful open-vocabulary capability to downstream tasks, directly involving VLMs in some tasks is non-trivial, e.g., the original VLMs are trained on non-masked images, and as a consequence, their performance naturally declines with the masked input images in video instance segmentation~\cite{liang2022open}. Additionally, the computational cost of running $N$ times VLMs' vision encoder for $N$ instance within a frame is impractical.

\subsection{Open-Vocabulary Segmentation}
Open vocabulary segmentation is proposed by ZS3Net~\cite{bucher2019zero}, which segments objects by their corresponding text description, which may contain categories not have been seen during the training stage. The mainstream works, such as ZSSeg~\cite{xu2021simple}, ZegFormer~\cite{ding2022decoupling}, and OVSeg~\cite{liang2022open}, utilize a two-stage framework to achieve open vocabulary segmentation, \textit{i.e.,} first extracting all potential proposals without classes from an image, then calculating the similarity between the visual features of proposals and text features of potential text descriptions to identify the categories. Recent concurrent work SAN~\cite{xu2023san} and DeOP~\cite{han2023DeOP} begin addressing the impractical computation cost of multiple passes of VLMs vision encoder in above methods, similar to the InstCLIP in our framework. However, compared to the above works, InstCLIP only needs a lightweight finetuning without changing VLM pre-trained weights and provides instance token for follow-by video-level instance association.

\section{Setting}

\subsection{Problem Formulation}
\label{definition}
Open-vocabulary Video Instance Segmentation (OpenVIS) aims to simultaneously segment, detect, and track open-world objects of arbitrary category based on the category name or corresponding text description in a video, regardless of whether the category has been seen during training. We are given a video consisting of $T$ frames, denoted as $\{F_t \in \mathbb{R}^{3\times H\times W}\}_{t=1}^{T}$, where $H$ and $W$ represent the height and width of each frame, respectively. Additionally, we have a set of category labels denoted as $\mathcal{C}$, which represents the possible categories of objects present in the video. Our objective in OpenVIS is to accurately predict all $N$ objects belonging to these categories within the video. Specifically, for each object $i$, its category label $ c^i \in \mathcal{C} $ and segmentation masks across the video $\textbf{m}^i_{p...q} \in \mathbb{R}^{H\times W\times (p-q)}$ need to be predicted, where $p\in [1,T]$ and $q\in[p,T]$ indicate its starting and ending frame index.

\subsection{Evaluation Benchmark}
\label{benchmark}
    To comprehensively evaluate the overall performance of the proposed OpenVIS, we introduce a novel evaluation benchmark. An ideal OpenVIS model should possess two essential properties, which form the two core focus of our evaluation: 1) open-world proposal ability to segment all possible instances within the video accurately and 2) zero-shot capability to correctly classify instances of arbitrary category. Moreover, we also further evaluate 3) the overall OpenVIS performance on both seen and unseen categories.
    \begin{itemize}
        \item \textbf{Open-world Property:} We leverage the exhaustively annotated UVO dataset \cite{wang2021uvo} to evaluate the open-world mask proposal ability. The UVO dataset provides an average of 13.52 instances annotated per video. Compared to the only 1.68 objects in YouTube-VIS, UVO is naturally a suitable dataset for measuring open-world property. 
    
    \item \textbf{Zero-shot Property:} We utilize the category-rich BURST dataset~\cite{athar2023burst} to evaluate the zero-shot instance classification property. The objects in BURST involve 482 categories, with 78 common categories from COCO~\cite{lin2014coco} and 404 uncommon categories, which can be regard as unseen categories. The uncommon-404 categories is an ideal dataset to measure the zero-shot property. Additionally, a latest dataset LVVIS~\cite{wang2023towards} novel set contains 555 unseen categories for our setting, so we also evaluate it.

    \item \textbf{Overall Property:} To further evaluate the overall performance on both seen and unseen categories, we also report the results on full BURST (482 categories).
    \end{itemize}
    Following \cite{yang2019video}, we utilize the Average Precision (i.e., AP) and Average Recall (i.e., AR) at the video level as the main metrics. Additionally, our OpenVIS model is only trained on YouTube-VIS (a widely-used VIS dataset comprising 40 categories). This ensures that the categories present in the training data are small-scale subsets of those found in the test data. More discussion and analysis of the evaluation benchmark can be found in \textit{Supplementary}.

\begin{figure*}[h]
    \centering
    \includegraphics[width=0.92\linewidth]{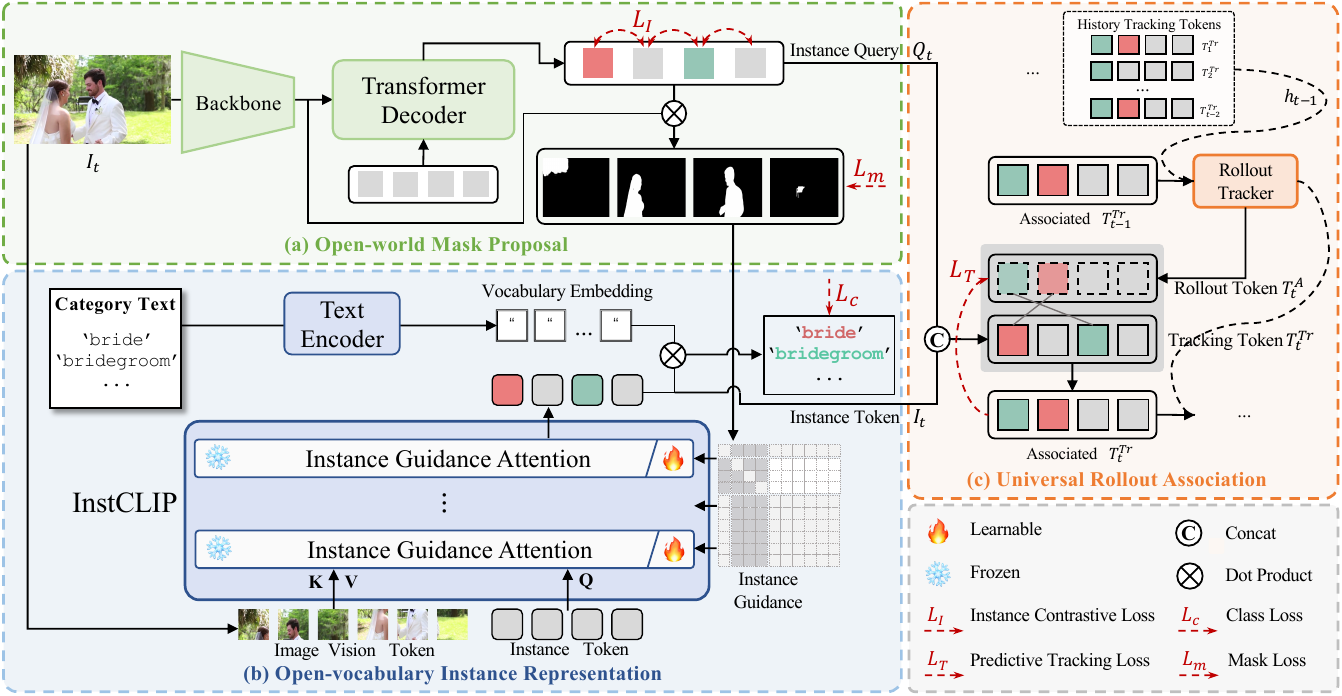}
    \vspace{-2mm}
    \caption{Overview of the proposed InstFormer framework for OpenVIS. (a) \textbf{Open-world Mask Proposal}: Generate class-agnostic instance masks with a query-based transformer, which is encouraged to propose all potential object instances. (b) \textbf{Open-vocabulary Instance Representation}: InstCLIP embeds open-vocabulary instance tokens using Instance Guidance Attention efficiently. These tokens enable open-vocabulary instance classification and provide robust open-vocabulary tracking capabilities. (c) \textbf{Universal Rollout Association}: Associate instances of any category across frames with the proposed universal rollout tracker, which is trained to predict the instance tracking tokens of the next frame, termed the rollout token.}
    \label{fig:pipeline}
    \vspace{-4mm}
\end{figure*}

\section{Method}

In this section, we detail how we bridge the gap between static, image-level VLMs and the video, instance-level demands of the OpenVIS task, leading to the InstFormer, a carefully designed framework tailored for OpenVIS that achieves open-vocabulary capabilities through lightweight fine-tuning on limited-category labeled data (as illustrated in Fig.~\ref{fig:pipeline}).

\subsection{Open-world Mask Proposal}
\label{sec:ow_mask_proposal}
An open-world proposed mask proposal network needs to propose as many distinct instances as possible to meet the flexible needs of open-world perception, as each instance has the possibility of being selected to identify by end user. 
To achieve this goal, we first adopted a query-based image segmentation model Mask2Former~\cite{cheng2022mask2former} as the mask proposal network, predicting $N$ class-agnostic masks $M_t=\{m_t^i\}_{i=1}^N \in [0,1]^{N\times H\times W }$ and their corresponding instance queries $Q_t=\{q_t^i\}_{i=1}^N\in \mathbb{R}^{N\times C}$ for each frame $F_t \in \mathbb{R}^{3\times H\times W}$ of a video:
\begin{equation}
    M_t, Q_t = \Psi (\Phi(F_t), Q^0),
\end{equation}
where $\Phi$ and $\Psi$ indicate the backbone and transformer decoder of the mask proposal network respectively. The ${Q^0}\in\mathbb{R}^{N\times C}$ denotes the $N$ learnable initial query embeddings. 
Despite the mask proposal network mentioned above can generate category-agnostic masks for all candidate instance, its training process on a dataset with a limited number of objects results in the redundant assignment of instance queries to the same instance. To ensure that these initial queries can perceive as many distinct instances as possible on given video, we introduce a contrastive instance margin loss to the open-world mask proposal network:
\begin{equation}
    \mathcal{L}_{SC} = \sum_{i=0}^{N} \sum_{j=0}^{N}\mathrm{max}(0, \mathrm{cos}(Q_t^i, Q_t^j)-\alpha ),
\end{equation}
where $\mathrm{cos}(\cdot, \cdot)$ refers to the cosine similarity ranging $[-1, 1]$, and $\alpha$ is the margin that determines how similar tokens should be penalized. This loss function will penalize instances that are excessively similar, thereby promoting diverse assignments of queries to distinct instances.

\subsection{Open-vocabulary Instance Representation}
\label{sec:ov_class}
Leveraging pre-trained VLMs like CLIP~\cite{radford2021learning} for zero-shot capabilities by directly inputting masked instance images is suboptimal and inefficient for real-time video tasks due to the domain gap between pre-training on natural images and testing on masked images, as well as the need to run CLIP vision encoder $N$ times per frame. The proposed InstCLIP efficiently represents each instance with an instance token, enabling open-vocabulary instance classification and providing strong open-vocabulary tracking capabilities. 
Specifically, InstCLIP is a Vision Transformer~\cite{dosovitskiy2020image} adapted from the pre-trained CLIP vision encoder, consisting of $L$ Instance Guidance Attention layers. We generate attention masks from mask proposals for Instance Guidance Attention to guide $N$ instance tokens to embed $N$ instances in a single forward pass through the encoder. Instance Guidance Attention takes as input the concatenated tokens $X_t^{l-1} \in \mathbb{R}^{1+N+P}$ from the previous attention layer and the guidance attention mask $\mathcal{M} \in \mathbb{R} ^ {(1+N+P) \times (1+N+P)}$:
\begin{align}
    X_t^l &= \mathrm{InstAttn}(X_t^{l-1}, \mathcal{M}) \\
          &\quad = \mathrm{softmax}(W^q X_t^{l-1} \cdot W^k X_t^{l-1} + \mathcal{M}) \cdot W^v X_t^{l-1}, \notag
\end{align}
where $W^q, W^k, W^v$ are weights of query, key, and value projection layer, respectively. $X_t^{l-1}$ consists of vision tokens $V_t \in \mathbb{R}^{\frac{H}{32} \times \frac{W}{32} \times C}$ from image patch embedding, $N$ initial instance tokens $I^l\in\mathbb{R}^{N\times C}$, and a register token $R^l\in\mathbb{R}^{1\times C}$. $P$ is the number of vision tokens.
The register token, inspired by \cite{darcet2023vision}, is a token permitted to attend to all vision tokens. It plays the role of collecting low-informative feature, which helps obtain cleaner attention maps from instance tokens to vision tokens.

\begin{figure}[t]
    \centering
    \includegraphics[width=1.0\linewidth]{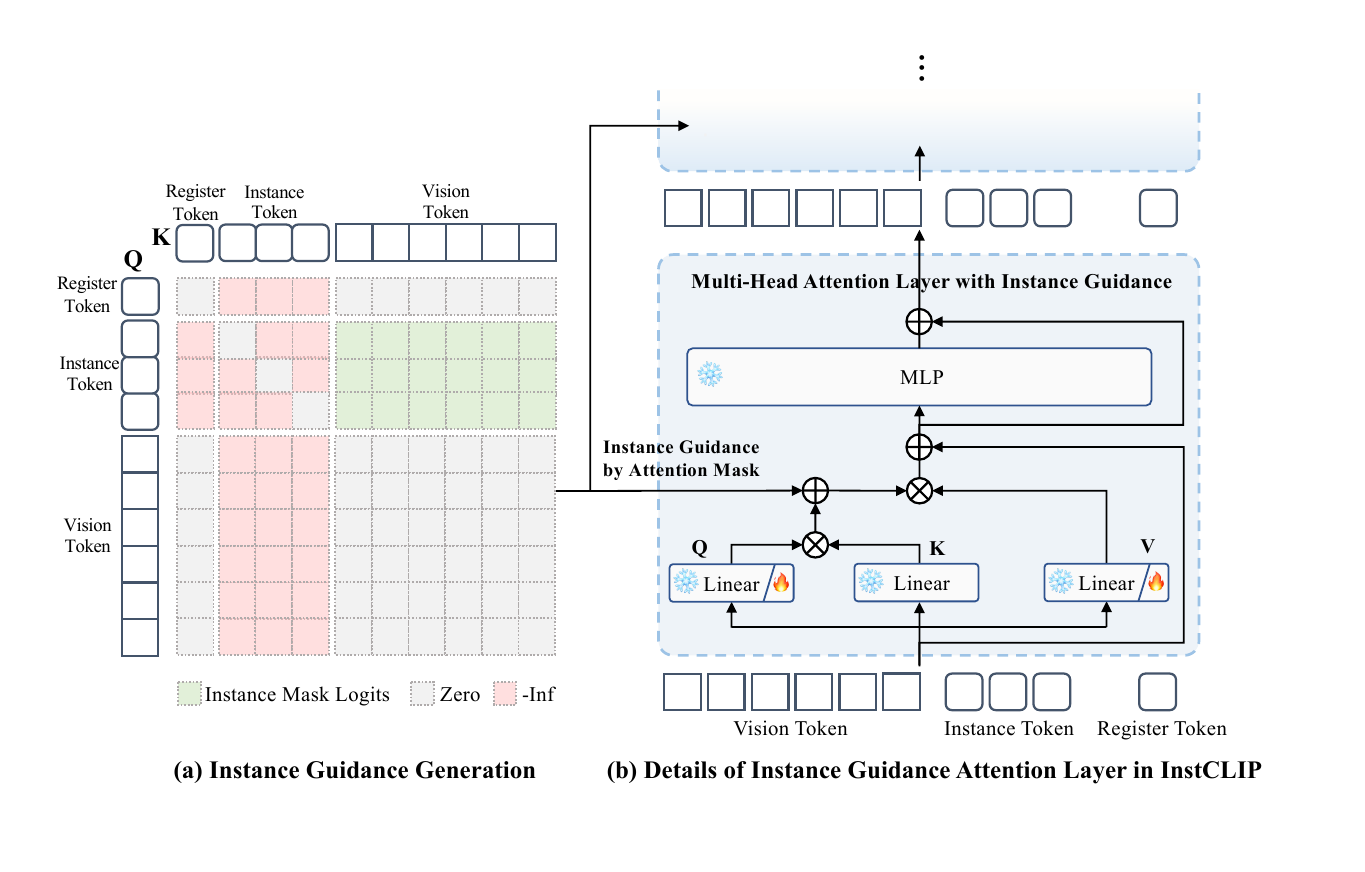} 
    \vspace{-4mm}
    \caption{The architecture of InstCLIP and the generation of the corresponding instance guidance mask.}
    \label{fig:instclip}
    \vspace{-4mm}
\end{figure}

The initial instance tokens $I^0$ and register token $R^0$ are learnable embeddings. The instance guidance $\mathcal{M}$, generated as illustrated in Fig.~\ref{fig:instclip}, directs instance tokens to attend different instance regions by acting as the attention mask in self-attention layers of the vision transformer. Instance tokens are independently guided to enhance attention to specific regions while suppressing attention to other regions based on the logits value of the instance masks. After $L$ instance guidance attention layers, these $N$ instance tokens aggregate CLIP features of $N$ instance.  So classification can be directly calculated by comparing them with vocabulary embeddings extracted by the CLIP text encoder: 
\begin{equation}
    C_t = \operatorname{argmax}( \operatorname{softmax}(I^L_t \cdot E^\top) ) \in \mathbb{K}^{N},
\end{equation}
where $E\in \mathbb{R}^{K\times C}$ is the vocabulary embeddings of $K$ categories. InstCLIP is designed with the principle of minimizing modifications from CLIP, to fully unleash the zero-shot capability of the pre-trained CLIP. Only the linear projections for the query and value of the attention layer are adjusted using the parameter-efficient fine-tuning approach LoRA~\cite{hu2021lora} during training, while almost parameters of CLIP remain frozen.

\subsection{Universal Rollout Association}
\label{sec:track}

To prevent trackers optimized on closed-set data from failing to generalize to open-vocabulary instance tracking, the proposed Universal Rollout Association fully leverages the open-vocabulary characteristics of instance tokens and transforms the tracking problem into predicting the next frame’s instance tracking token for training.

\paragraph{\textbf{Instance Tracking Tokens}.}
We form universal instance tracking tokens $T^{Tr}_t$ by combining the instance tokens $I_t$ from InstCLIP with the instance queries $Q_t$ from the proposal network to handle open-vocabulary tracking: $T^{Tr}_t = \mathrm{Concat}(I_t, Q_t) \in \mathbb{R}^{N\times 2C}$. These tokens and queries are naturally aligned in a token/query-based architecture.
The former, which leverages CLIP features with zero-shot capabilities, proves particularly effective for tracking open-vocabulary instances in subsequent experiments. The latter, characterized by its ability to generate class-agnostic mask proposals, has also been demonstrated in MinVIS~\cite{huang2022minvis} to distinguish instances between frames.

\paragraph{\textbf{Rollout Association}.} And to prevent the tracker from being constrained by fixed-category object data, we reframe the tracking problem by training the tracker to predict the instance tracking token for the next frame. When associating instances in frame-t with those from previous frames, the rollout tracker predicts the instance tracking token for frame-t based on the tokens from history frames. This predicted instance tracking token, referred to as the rollout association token, is denoted as  $T^A_t \in \mathbb{R}^{N \times C}$:
\begin{equation}
    T^A_t=\mathcal{R}(T^{Tr}_{1\rightarrow t-1})=\mathrm{RNN} (T^{Tr}_{t-1}, h_{t-1}),
\end{equation}
where $\mathcal{R}$ denotes the concise yet effective rollout tracker implemented by a single RNN layer and $h_{t-1}$ is the hidden state of RNN remaining instance temporal information. Finally, by comparing the rollout association token for frame t with the actual instance tracking token using Hungarian matching on the similarity score  $S_{ij} = \mathrm{cos}(T^A_t, T^{Tr}_t)$, the instance association for frame-t can be completed.
The rollout tracker is trained with the loss:
\begin{equation}
    \mathcal{L}_{T} = \sum_{i=1}^{N}\sum_{j=1}^{N} \mathrm{CE}(\mathrm{cos} (T^{A}_{t}(i), T^{Tr}_{t}(j)),  \one{i=j} ),
\end{equation}
where $\one{i=j} \in \{0, 1\}$ is an indicator function evaluating to $1$ if $i=j$. This process is independent of categories, allowing the rollout tracker to handle open-vocabulary instances. Meanwhile, the incorporation of historical information in this history-aware tracker enhances robustness compared to tracking based solely on the previous frame. 

\section{Experiment}

\begin{table*}[]
\centering
\begin{tabular}{lcccccc}
\toprule
\multirow{2}{*}{Method} & \multirow{2}{*}{OV} & Training \quad & \multicolumn{3}{c}{BURST} & \multirow{2}{*}{LVVIS$_{novel}$} \\ \cline{4-6}
                        &  & Categories \quad & \textit{All} & \textit{Common} & \textit{Uncommon} &  \\
\midrule
\multicolumn{7}{c}{{\color[RGB]{128,128,128} \textit{Fully-supervised}}}    \\
\hline
MRCNN~{\cite{maskrcnn}}-BoxTracker  & $\times$  & L-1203 & 1.4 \scriptsize${\color{gray}0\%}$ & 3.0 \scriptsize${\color{gray}0\%}$  & 0.9 \scriptsize${\color{gray}0\%}$  & - \\
MRCNN~\cite{maskrcnn}-STCNTracker &  $\times$ & L-1203 & 0.9 \scriptsize${\color{gray}0\%}$ & 0.7 \scriptsize${\color{gray}0\%}$  & 0.6 \scriptsize${\color{gray}0\%}$  & -  \\
MinVIS~\cite{huang2022minvis}               & $\times$  & CB-482 & 1.4 \scriptsize${\color{gray}0\%}$ & 5.5 \scriptsize${\color{gray}0\%}$  & 0.5 \scriptsize${\color{gray}0\%}$  & -  \\
\midrule
\multicolumn{7}{c}{{\color[RGB]{128,128,128} \textit{Open-vocabulary}}}    \\
\hline
Detic~\cite{zhou2022detecting}-SORT~\cite{bewley2016simple}              & \checkmark & L-866  & 1.9 \scriptsize${\color{gray}15\%}$ & 1.8 \scriptsize${\color{gray}0\%}$   & 2.5 \scriptsize${\color{gray}18\%}$  & 3.4 \scriptsize${\color{gray}100\%}$  \\
Detic~\cite{zhou2022detecting}-OWTB~\cite{liu2022opening}              & \checkmark & L-866  & 2.7 \scriptsize${\color{gray}15\%}$ & 2.8 \scriptsize${\color{gray}0\%}$   & 1.8 \scriptsize${\color{gray}18\%}$  & 4.2 \scriptsize${\color{gray}100\%}$ \\
OV2Seg~\cite{wang2023towards}                  & \checkmark & L-866  & 3.7 \scriptsize${\color{gray}15\%}$ & 3.9 \scriptsize${\color{gray}0\%}$   & 2.4 \scriptsize${\color{gray}18\%}$  & 11.9 \scriptsize${\color{gray}100\%}$  \\
InstFormer (Ours)                 & \checkmark & CY-103 & \textbf{4.2} \scriptsize${\color{gray}84\%}$ & \textbf{7.4} \scriptsize${\color{gray}0\%}$   & \textbf{3.5} \scriptsize${\color{gray}96\%}$  & \textbf{12.2} \scriptsize${\color{gray}100\%}$ \\   
\bottomrule
\end{tabular}
\vspace{-2mm}
\caption{Overall OpenVIS performance and zero-shot property comparison with baselines on BURST and LVVIS with AP metric. OV indicates whether the method has the ability to handle the open-vocabulary setting. The Training Categories column shows the training dataset and the number of categories involved. The {\color{gray} gray\%} represents the proportion of novel categories during inference for each approach and setting (higher means more challenging).}
\label{tab:comp_ov}
\vspace{-5mm}
\end{table*}

\subsection{Implementation Details}

\paragraph{\textbf{Model Architecture}.} We regard a COCO~\cite{lin2014coco}-pretrained Mask2Former~\cite{cheng2022mask2former} as our mask proposal network. By default, the transformer decoder has 100 queries, with a dimension of 256 for the query embedding and transformer decoder. For InstCLIP, we select a ViT-B/32 of CLIP~\cite{radford2021learning} as  foundation vision transformer. The number of instance tokens of InstCLIP is also set to 100, aligning with the 100 instance queries of the mask proposal network. We initialize the instance tokens and register token using CLIP's learned class tokens. The text encoder is a 12-layer transformer, the same as that in CLIP. For input prompts, we ensemble 14 prompts (e.g., ``a photo of a \{category name\}'') from \cite{liang2022open} to boost zero-shot classiﬁcation ability. 

\paragraph{\textbf{Training}.} InstFormer is trained using a two-stage approach and CLIP weights are frozen during the entire training. In first stage, the open-world mask proposal network and InstCLIP (LoRA adapter) are trained for 6k iterations with $\mathcal{L}_{I}$ and instance segmentation loss. Subsequently, we train the rollout tracker in second stage, with all other weights frozen, using $\mathcal{L}_{T}$ for an additional 600 iterations. The whole training is done on 8 V100 GPUs for 3 hours.

\paragraph{\textbf{Baselines}.} 
To better assess the performance of the proposed InstFormer framework, we introduce several baselines for comparison, as shown in Tab.~\ref{tab:comp_ov}. For fully-supervised methods, we provide STCN Tracker~\cite{athar2023burst}, Box Tracker~\cite{athar2023burst} and MinVIS~\cite{huang2022minvis}. Both the first two methods ultilize the Mask-RCNN~\cite{maskrcnn} but with different tracking strategies as in \cite{athar2023burst}. MinVIS~\cite{huang2022minvis} is an advanced VIS model in fully-supervised VIS task. All of them are trained on full BURST dataset with all 482 categories. For open-vocabulary methods, we employ three approaches: Detic-SORT, Detic-OWTB, and OV2Seg~\cite{wang2023towards}. The first two methods utilize the open-vocabulary detector Detic~\cite{zhou2022detecting}, paired with the classical multi-object tracker SORT~\cite{bewley2016simple} and the state-of-the-art open-world tracker OWTB~\cite{liu2022opening}, respectively. OV2Seg introduces the CLIP text encoder and a momentum-updated query for tracking, also achieving open-vocabulary video instance segmentation. 
These baselines may be trained with different datasets. We provide their corresponding training datasets and category numbers in Tab.~\ref{tab:comp_ov}. L-1203 represents the entire LVIS~\cite{gupta2019lvis} dataset with all 1203 categories. L-866 indicates the LVIS subset with 866 frequent categories. B and Y denote BURST~\cite{athar2023burst} with 482 categories and YouTube-VIS~\cite{yang2019video} with 40 categories, respectively. C is the COCO~\cite{lin2014coco} dataset with 80 categories, which may be used to pretrain some modules. 
And for clarity, we provide the proportion of novel categories during inference for each approach and setting in the table in {\color{gray} gray\%} (higher means more challenging).

\subsection{Main Results}

\begin{table}[]
\setlength\tabcolsep{2pt}
\centering
\begin{tabular}{lcccc}
\toprule
               & \scriptsize{Training Data} & AP   & AP$_{c1}$ & $AR_{100}$ \\ 
\midrule
MTRCNN\scriptsize{~\cite{yang2019video}} & \scriptsize{YouTubeVIS}    & 7.6  & -                  & 9.3     \\
MTRCNN\scriptsize{~\cite{yang2019video}} & \scriptsize{UVO}           & 11.2 & -                  & 17.4    \\
TAM\scriptsize{~\cite{yang2023track}} & \scriptsize{SA-1B} & -    & 1.7                & 24.1    \\
SAM-PT\scriptsize{~\cite{rajivc2023segment}} & \scriptsize{SA-1B} & -    & 6.7                & \textbf{28.8}    \\
InstFormer\scriptsize{~(Ours)}    & \scriptsize{YouTubeVIS}    & \textbf{16.7} & \textbf{7.2}                & 24.7    \\ 
\bottomrule
\end{tabular}
\vspace{-3mm}
\caption{Comparison of open-world instance proposal property on UVO. AP$_{c1}$ indicates the class-agnostic AP.}
\label{tab:ow}
\vspace{-5mm}
\end{table}

\paragraph{\textbf{Overall Performance}.}
We evaluate the overall performance of the InstFormer on the BURST validation set. Since the mask proposal network is pre-trained on COCO and InstFormer is trained on YouTube-VIS, so there are 103 categories have been seen during training. As illustrated in Tab.~\ref{tab:comp_ov}, our proposed InstFormer framework outperforming fully-supervised baselines by a large margin (AP from 1.4 to 4.2). And despite InstFormer seen fewer categories compared to other open-vocabulary baselines, it still achieved state-of-the-art OpenVIS performance (BURST 4.2 AP, LVVIS$_{novel}$ 12.2 AP), demonstrating InstFormer achieves obvious advantages over other methods. Qualitative results can be found in Fig.~\ref{fig:teaser} and \textit{Supplementary}.

\paragraph{\textbf{Zero-shot Instance Classification}.}

To measure zero-shot instance classification property, we report the results of BURST-uncommon with 404 categories and LVVIS-novel with 555 categories in Tab.~\ref{tab:comp_ov}. 
Specifically, for BURST-uncommon, where 96\% categories are novel to us, we achieve a 45\% (AP from 2.4 to 3.5) improvement over the OV2Seg. 
For LVVIS-novel, we also achieve the best performance even InstFomer only seen 103 categories, which the compared methods like OV2Seg have seen 866 categories. This demonstrates that InstCLIP successfully maintains the zero-shot capability of the pre-trained CLIP model.

\paragraph{\textbf{Open-world Instance Proposal}.}
In this section, we evaluate the performance of the open-world mask proposal, which is a critical component for achieving OpenVIS, using the extensively annotated UVO dataset. As reported in Tab.~\ref{tab:ow}, our mask proposal network, trained solely on the YouTube-VIS dataset with contrastive instance margin loss, outperforming the baseline method~\cite{yang2019video} traind on YouTube-VIS and even on UVO itself. Compared with the most advanced mask proposal approaches empowered by the Segment Anything Model(SAM)~\cite{kirillov2023segment} trained with the extensive dataset SA-1B~\cite{kirillov2023segment}, our mask proposal network can also achieve comparable performance.

\begin{table}[]
\setlength\tabcolsep{3pt}
\centering
\begin{tabular}{llllll}
\toprule
                                   & OV         & AP         & AP$_{50}$           & AP$_{75}$             & AR$_{10}$              \\ 
\midrule
\multicolumn{6}{c}{{\color[RGB]{128,128,128} \textit{Fully-supervised}}}    \\
\hline
MaskTrack\tiny{~\cite{yang2019video}}                          & $\times$                 & 30.3                        & 51.1                        & 32.6                        & 35.5                        \\
SipMask\tiny{~\cite{cao2020sipmask}}                          & $\times$                 & 33.7                        & 54.1                        & 35.8                        & 40.1                        \\
CrossVIS\tiny{~\cite{yang2021crossover}}                          & $\times$                 & 36.3                        & 56.4                        & 38.9                        & 40.7                        \\
VISOLO\tiny{~\cite{han2022visolo}}                          & $\times$                 & 38.6                        & 56.3                        & 43.7                        & 42.5                        \\
MinVIS\tiny{~\cite{huang2022minvis}}                          & $\times$                 & 47.4                        & 69.0                        & 52.1                        & 55.7                        \\
IDOL\tiny{~\cite{wu2022defense}}                               & $\times$                 & 49.5                        & 74.0                        & 52.9                        & 58.1                        \\
GenVIS\tiny{~\cite{heo2023generalized}}                             & $\times$                 & 50.0                        & 71.5                        & 54.6                        & 59.7                        \\
DVIS\tiny{~\cite{zhang2023dvis}}                               & $\times$                 & 51.2                        & 73.8                        & 57.1                        & 59.3                        \\
\underline{MinVIS-CLIP}            & \checkmark           & \underline{30.6}            & \underline{51.2}            & \underline{32.0}            & \underline{40.7}    \\
InstFormer (Ours)                            & \checkmark               & \textbf{51.8}               & \textbf{75.6}               & \textbf{57.2}               & \textbf{60.0}               \\
\midrule
\multicolumn{6}{c}{{\color[RGB]{128,128,128} \textit{Open-vocabulary}}}    \\
\hline
{\color{gray} Detic-SORT}  & {\color{gray} \checkmark} & {\color{gray} 14.6} & {\color{gray} -}    & {\color{gray} -}   & {\color{gray} -}    \\
{\color{gray} Detic-OWTB}  & {\color{gray} \checkmark} & {\color{gray} 17.9} & {\color{gray} -}    & {\color{gray} -}   & {\color{gray} -}    \\
{\color{gray} Ov2seg}      & {\color{gray} \checkmark} & {\color{gray} 27.2} & {\color{gray} -}    & {\color{gray} -}   & {\color{gray} -}    \\ 
\bottomrule
\end{tabular}
\vspace{-3mm}
\caption{Performance comparison in the fully-supervised VIS on YouTube-VIS.}
\label{tab:vis}
\vspace{-7mm}
\end{table}

% \vspace{-2mm}
\paragraph{\textbf{Fully-supervised VIS}.}
An ideal OpenVIS model should handle both open set and closed set problems well. So we also evaluate the performance of the proposed model on YouTube-VIS. As shown in Tab.~\ref{tab:vis}, notably, the proposed InstFormer for open-vocabulary achieves top-tier performance in the fully-supervised VIS area, whether compared to other OpenVIS baselines or even the fully-supervised approaches.
However, as InstFormer is trained with YouTubeVIS while other open-vocabulary baselines are not, we also implement an open-vocabulary baseline trained with YouTube-VIS, namely the MinVIS-CLIP, for a fair comparison. It replaces the closed-set classification head of the MinVIS (trained on YouTube-VIS) with a frozen CLIP for open-vocabulary ability. MinVIS-CLIP exactly is the starting-point baseline for our framework. Experiments show that InstFormer outperforms MinVIS-CLIP by a significant margin of 21.2 AP, demonstrating that the methods we proposed for open-vocabulary are also beneficial for regular fully-supervised VIS tasks.

\subsection{Ablation Study}
In this section, we conduct ablation study on key designs of our framework to demonstrate their effectiveness. In the experiments of non-tracker components, to avoid performance changes caused by trackers, we default to using Hungarian matching with the instance query for association.

\begin{table}[]
\setlength\tabcolsep{3pt}
\centering
\begin{tabular}{llccc}
\toprule
~ & \multicolumn{1}{c}{}       & BURST & YouTube-VIS & UVO  \\
\midrule
\footnotesize \textcolor{darkF7E0D5}{1} & MinVIS-CLIP                 & 2.1   & 30.6        & 9.0  \\
\footnotesize \textcolor{darkF7E0D5}{2} & + InstCLIP                  & 3.3   & 48.6        & 13.2 \\
\footnotesize \textcolor{darkF7E0D5}{3} & + $\mathcal{L}_{SC}$       & 3.5   & 48.5        & 15.8 \\
\footnotesize \textcolor{darkF7E0D5}{4} & + InstCLIP Token     & 3.9   & 50.2        & 16.1 \\
\footnotesize \textcolor{darkF7E0D5}{5} & + Rollout Tracker & \textbf{4.2}   & \textbf{51.8}        & \textbf{16.7} \\ 
\bottomrule
\end{tabular}
\vspace{-3mm}
\caption{Ablation study of InstFormer on diverse datasets.}
\label{tab:abl}
\vspace{-3mm}
\end{table}

\paragraph{\textbf{Effectiveness of InstCLIP}.}

In Tab.~\ref{tab:abl}, comparing Line {\textcolor{darkF7E0D5}{1}} and Line {\textcolor{darkF7E0D5}{2}}, InstCLIP demonstrates significant improvements over the MinVIS-CLIP baseline, where the masked image is directly input into CLIP for zero-shot classification. Specifically, on BURST, we observe an increase in AP from 2.1 to 3.3 (57\% improvement). Similarly, on UVO, the AP rises from 9.0 to 13.1 (46\% increase). Notably, for YouTube-VIS, there is a remarkable gain of 18 AP (58\% improvement). 
This shows that InstFormer retains most of CLIP’s zero-shot capabilities while optimizing for YouTubeVIS domain, offering an effective solution for scenarios requiring both extreme domain performance and generalization.

\begin{table}[]
\setlength\tabcolsep{3pt}
\centering
\begin{tabular}{llcccc}
\toprule
    &       & AP$_{All}$ & AP$_{Com}$ & AP$_{Uncom}$ & Once \\
\midrule
\footnotesize \textcolor{darkF7E0D5}{1} & $N$ times CLIP       & 2.11    & 3.58       & 1.81         & $\times$       \\
\footnotesize \textcolor{darkF7E0D5}{2} & + Instance Token    & 1.09    & 0.87       & 1.14         & \checkmark     \\
\footnotesize \textcolor{darkF7E0D5}{3} & + Binary Guidance & 1.70    & 2.13       & 1.62         & \checkmark     \\
\footnotesize \textcolor{darkF7E0D5}{4} & + Designed Guidace & 3.28    & 6.88       & 2.52         & \checkmark     \\
\footnotesize \textcolor{darkF7E0D5}{5} & + Register Token   & \textbf{3.87} & \textbf{7.01} & \textbf{3.22} & \checkmark \\
\bottomrule
\end{tabular}
\vspace{-3mm}
\caption{Ablation study of InstCLIP on BURST.}
\label{tab:abl_instclip}
\vspace{-6mm}
\end{table}

\paragraph{\textbf{Instance Tokens for Association}.}
InstCLIP's instance tokens not only achieve open-vocabulary classification effectively and efficiently, as illustrated in Tab.~\ref{tab:abl} Line {\textcolor{darkF7E0D5}{4}} and Tab.~\ref{tab:abl_tracker}, but also aid in tracking instances of any vocabulary.

\paragraph{\textbf{Key design of InstCLIP}.}
In this part, we ablate the key design of InstCLIP including instance guidance mask, instance tokens, and register tokens. In Tab.~\ref{tab:abl_instclip}, directly introducing $N$ instance tokens into CLIP to enable CLIP to classify $N$ instances in a single-forward doesn't work well, as instance representations cannot aggregate into tokens without specific guidance (Line \textcolor{darkF7E0D5}{2}).
Masking the background region for each instance with the binary instance mask from mask proposal network allows instance token to know what should attend (Line \textcolor{darkF7E0D5}{3}). Line \textcolor{darkF7E0D5}{4} reveals that the effectiveness of InstCLIP hinges on the generated Instance Attention Mask. The register token, specifically designed for collecting low-informative features, indeed assists InstCLIP in obtaining superior instance tokens and vision tokens (Line \textcolor{darkF7E0D5}{5}).

\paragraph{\textbf{Contrastive Instance Margin Loss}.}

We study the effect of the contrastive margin loss to open-world mask proposal on UVO. As shown in Tab.~\ref{tab:abl} Line {\textcolor{darkF7E0D5}{3}} and Tab.~\ref{tab:abl_ow_loss} of Supp, introducing the contrastive instance margin loss encourages the mask proposal network provide more distinct instances, thereby improving both AP and AR. More details see Supp.

\paragraph{\textbf{Effectiveness of Rollout Association}.}
\label{sec:abl_tracker}
Tab.~\ref{tab:abl_tracker} ablates the key components of the rollout association. 
Given the instance token provides a richer open-vocabulary tracking feature, the predictive tracking loss-driven rollout tracker achieves a 10.3\% improvement in AP performance in the OpenVIS setting. Additionally, it provides a 3.2\% boost in fully-supervised VIS tasks, demonstrating the effectiveness of rollout association in normal fully-supervised tracking instances. RNN offers historical information offered by the hidden state, aiding in handling object occlusion and reappearance issues. We also replace the RNN layer with a linear layer or a two-layer MLP with the larger capacity.

\begin{table}[]
    \centering
    \begin{tabular}{ccccc}
    \toprule
    \begin{tabular}[c]{@{}c@{}}Instance\\ Token\end{tabular} & \begin{tabular}[c]{@{}c@{}}Rollout\\ Tracker\end{tabular} & $\mathcal{L}_{TC}$ & BURST & YouTube-VIS \\ 
    \midrule
    $\times$ & $\times$      & $\times$ & 3.5 & 48.5 \\
    \checkmark & $\times$      & $\times$ & 3.9 & 50.2 \\
    $\times$ & RNN    & \checkmark & 3.6 & 49.9 \\
    \checkmark & RNN    & \checkmark & \textbf{4.2} & \textbf{51.8} \\
    \checkmark & Linear & \checkmark & 3.7 & 49.1 \\
    \checkmark & MLP    & \checkmark & 3.4 & 49.7 \\
    \bottomrule
    \end{tabular}
    \vspace{-3mm}
    \caption{Ablation study of Rollout Association in both OpenVIS (on BURST) and fully-supervised VIS (on YouTube-VIS).}
    \vspace{-6mm}
    \label{tab:abl_tracker}
\end{table}

% \vspace{-2mm}
\section{Conclusion}
\vspace{-2mm}
In this paper, we propose InstFormer, a carefully designed framework tailored for the OpenVIS task that achieves powerful open-vocabulary capabilities through lightweight fine-tuning on a limited-category data, to bridge the gap between static, image-level VLMs and the video, instance-level demands of the task.
In the InstFormer framework, a mask proposal network encouraged to propose all potential object instances with the margin instance contrastive loss is proposed. 
Then we propose the InstCLIP to embed open-vocabulary instance tokens efficiently, which enable open-vocabulary instance classification and provide robust open-vocabulary tracking
Furthermore, we propose the universal rollout association where the rollout tracker is trained to predict instance tokens of the next frame to achieve universal tracking.
Last but not least, we propose a comprehensive evaluation benchmark to facilitate further research in this emerging task. Through extensive experiments, we demonstrate the proposed InstFormer achieves both state-of-the-art capabilities in OpenVIS and competitive performance in fully supervised VIS.

\bibliography{aaai25}

\clearpage
\clearpage
\setcounter{page}{1}

\begin{figure}[h]
    \centering
    \includegraphics[width=0.9\linewidth]{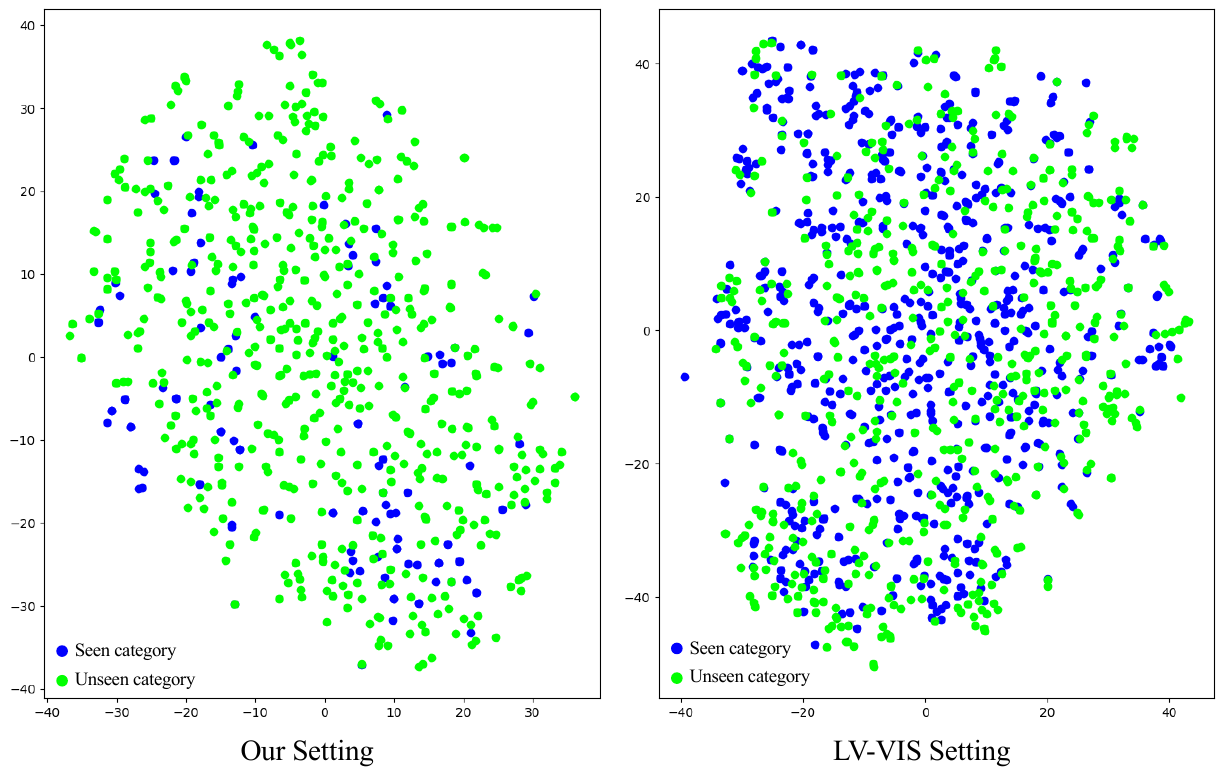}
    \caption{Comparison of categories distribution space. Category features are extracted using the text encoder of CLIP. The figure is presented after applying the t-SNE dimensionality reduction.}
    \label{fig:benchmark}
\end{figure}

\paragraph{A. Evaluation Benchmark}
In our evaluation benchmark, the model is trained on the YouTube-VIS dataset with 40 categories and is tasked with inference on BURST, including an additional 404 unseen categories from the dataset. 
Given the COCO pretrained mask proposal network, the model has encountered a total of 103 categories during training. 
There is concurrent work, LV-VIS~\cite{wang2023towards}, which also aims to establish a benchmark for open-vocabulary video instance segmentation. It trains the model on 866 categories from LVIS datasets and performs inference on LV-VIS with an additional 555 novel categories. 
While LV-VIS contributes to the community, it's noteworthy that our setting is more challenging, enabling the evaluation of open-vocabulary capabilities. As illustrated in Fig.~\ref{fig:benchmark}, our setting involves exposure to a limited number of seen categories (blue) in the training data, while the unseen categories (green) for inference are numerous and distinctly different from the seen categories. In contrast, the LV-VIS setting has seen categories occupying most of the space and being close to the unseen categories. 

\paragraph{B. Ablation of Contrastive Instance Margin}
Introducing the contrastive instance margin loss encourages the mask proposal network to provide more distinct instances. However, simply pushing all instance queries apart does not achieve this effect. It’s crucial to control the hyperparameter margin  $\alpha$  so that instance queries are only pushed apart when they are too similar. Empirically, the margin  $\alpha$  is set to 0.8.

\begin{table}[h]
    \centering
    \begin{tabular}{c|ccccc}
    \toprule
    Margin $\alpha$ & -1  & 0    & 0.5  & 0.8  & 1   \\
    \midrule
    $AR_{100}$ & 21.0 & 21.1 & 22.5 & \textbf{22.7} & 21.9 \\
    \bottomrule
    \end{tabular}
    \captionof{table}{Impact of similar instance margin $\alpha$ on UVO.}
    \label{tab:abl_ow_loss}
\end{table}

\paragraph{C. Ablation of Tuning Weight}
We also ablate the impact of tuning different pretrained weights of CLIP. As shown in Tab.~\ref{tab:abl_lora}, tuning the query and value projectors in the attention layer leads to a more favorable trade-off performance for both common and uncommon categories.

\begin{table}[h]
\centering
\begin{tabular}{lccc}
\toprule
\multicolumn{1}{c}{LoRA} & AP$_{All}$ & AP$_{Com}$ & AP$_{Uncom}$  \\
\midrule
FFN                 & 3.5   & 6.5 & 2.9   \\
Proj.~Q,V            & \textbf{3.9}   & \textbf{7.4} & 3.1   \\
Proj.~Q,K,V           & 3.6   & 6.5 & 3.0   \\
FNN + Proj.~Q,K,V     & 3.7   & 5.8 & \textbf{3.3}   \\
\bottomrule
\end{tabular}
\caption{Impact of inserting LoRA into different layers of pre-trained CLIP, on BURST.}
\label{tab:abl_lora}
\end{table}

\begin{figure}[h]
    \includegraphics[width=0.99\linewidth]{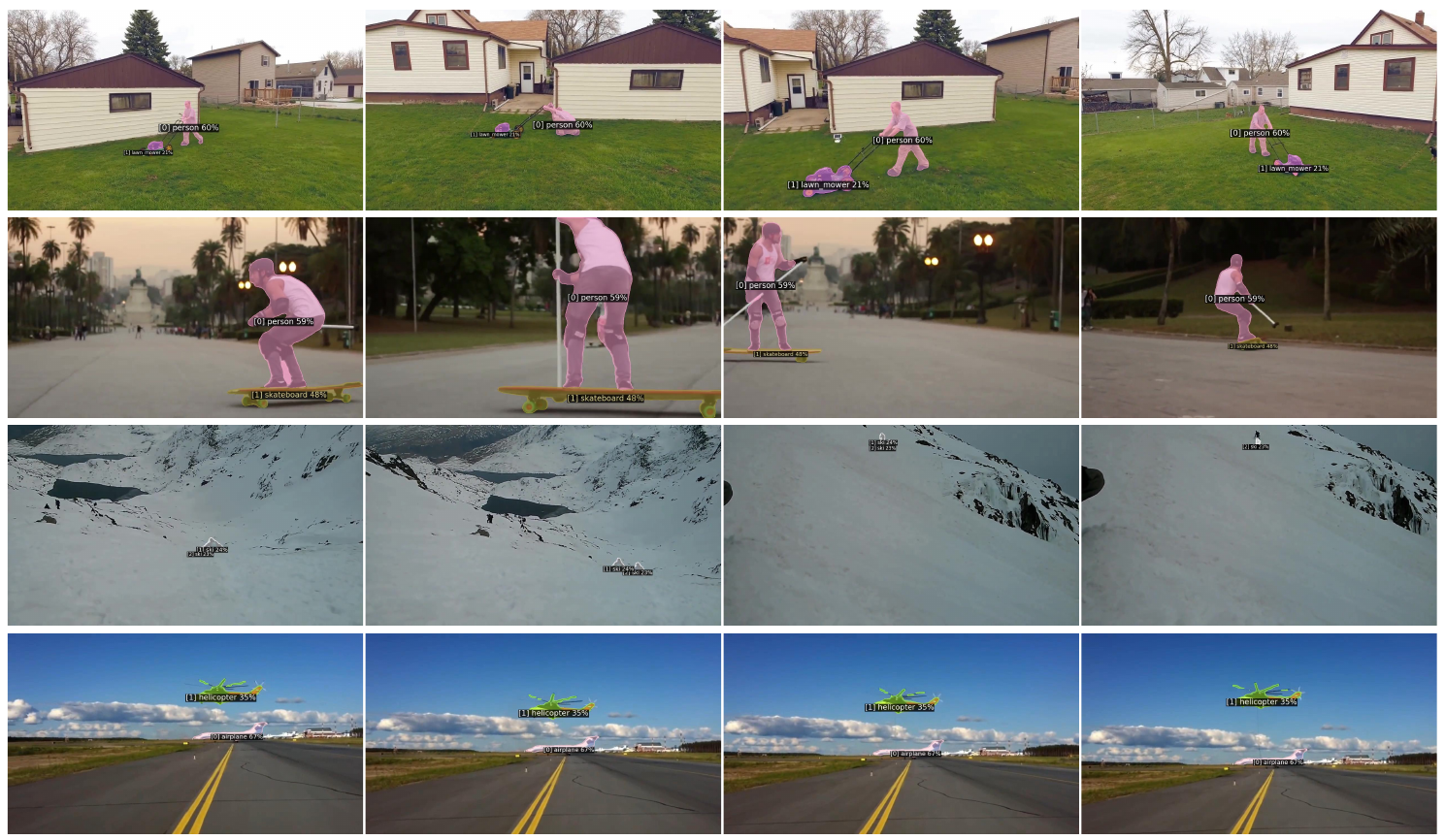}
    \caption{Qualitative results on BURST. The proposed InstFormer successfully recognizes uncommon categories such as ``lawn\_mower'' and even demonstrates the ability to distinguish between ``skateboard'' and ``ski'', as well as ``helicopter'' and ``airplane''.}
    \label{fig:qualitative}
\end{figure}

\paragraph{D. Qualitative Results}
We present qualitative results from the BURST evaluation, where inference involves 482 categories in Fig.~\ref{fig:qualitative}. The proposed InstFormer successfully recognizes uncommon categories such as ``lawn\_mower'' and even demonstrates the ability to distinguish between ``skateboard'' and ``ski'', as well as ``helicopter'' and ``airplane''.

\end{document}